\documentclass[10pt,twocolumn,letterpaper]{article}

\usepackage{iccv}
\usepackage{times}
\usepackage{epsfig}
\usepackage{graphicx}
\usepackage{amsmath}
\usepackage{amssymb}
\usepackage{authblk}
\usepackage{booktabs}
\usepackage{comment}
\usepackage[utf8]{inputenc}
\usepackage{array}
\usepackage{makecell}

\renewcommand\Affilfont{\fontsize{7.7}{14.4}\selectfont}

\usepackage[hang,flushmargin]{footmisc}
\makeatletter
\renewcommand\AB@affilsepx{  \hspace{1 mm}  \protect\Affilfont}
\makeatother

\iccvfinalcopy 
\def\iccvPaperID{****} 

\ificcvfinal\fi

\begin{document}

\title{AIM 2019 Challenge on Image Demoireing: Dataset and Study}

\author{
Shanxin Yuan$^1$ \hspace*{4mm}
Radu Timofte$^2$ \hspace*{4mm}
Gregory Slabaugh$^1$ \hspace*{4mm}
Ale\v{s} Leonardis$^1$ 
}
\affil{\large
$^1$Huawei Noah's Ark Lab \hspace*{4mm}
$^2$ETH Z\"{u}rich, Switzerland
}

\maketitle
\ificcvfinal\thispagestyle{empty}\fi

\begin{abstract}
This paper introduces a novel dataset, called \emph{LCDMoire}, which was created for the first-ever image demoireing challenge that was part of the \emph{Advances in Image Manipulation} (AIM) workshop, held in conjunction with ICCV 2019. The dataset comprises 10,200 synthetically generated image pairs (consisting of an image degraded by moire and a clean ground truth image). In addition to describing the dataset and its creation, this paper also reviews the challenge tracks, competition, and results, the latter summarizing the current state-of-the-art on this dataset.  
\end{abstract}

\section{Introduction}
\label{report:into}

Moire effects appear when two repetitive patterns interfere with each other. In the case of digital photography of screens, moire patterns occur when the screen's subpixel layout interferes with the camera's color filter array (CFA).  The moire pattern is detrimental to image quality, resulting in repetitive banding artifacts obscuring detail and corrupting the appearance of the image. Moire artifacts vary in both frequency and amplitude, but even subtle moire patterns can diminish the fidelity and perceptual quality of an image.

To address this issue, demoire methods \cite{sun2018moire, liu2018demoir, gao2019moire} seek to remove moire patterns to reveal an underlying clean image. Example-based image demoireing relies on prior examples in the form of moire and moire-free image pairs. In this context, demoireing can be cast as an image restoration problem addressable through supervised machine learning, in particular, deep learning. However, to train and evaluate a deep demoire network, it is necessary to have sufficient training data. In this paper, we propose a new synthetic dataset called \emph{LCDMoire} consisting of 10,200 synthetically generated image pairs (moire image, and clean ground truth). This dataset was created as part of the demoire challenge~\cite{AIM19demoireMethods} that was part of the \emph{Advances in Image Manipulation} (AIM\footnote{\url{http://www.vision.ee.ethz.ch/aim19/}}) workshop, held in conjunction with ICCV 2019.

This paper makes two contributions. First, we introduce the LCDMoire dataset, and describe it in terms of its composition and how it was created. This dataset is a new resource to the computer vision / image processing community, and we hope it inspires future work on the demoire problem and is used to benchmark algorithms. Second, we present the demoire results achieved at the workshop, providing the current state-of-the-art performance on this dataset. We report results using a selection of image quality measures. 

The remainder of the paper is structured as follows.
Section 2 introduces the LCDMoire dataset.
Section 3 reviews the AIM 2019 demoire competition tracks, competition phases, and the challenge results. 
Section 4 introduces the image quality assessment (IQA) measures,
Section 5 discusses related datasets.
Section 6 presents the methods from the workshop, and
Section 7 concludes the paper.

\section{LCDMoire dataset}
\label{study:dataset}

\begin{figure*}
	\scriptsize
	\centering

		\begin{tabular}{cc}
			\includegraphics[width=0.46\textwidth]{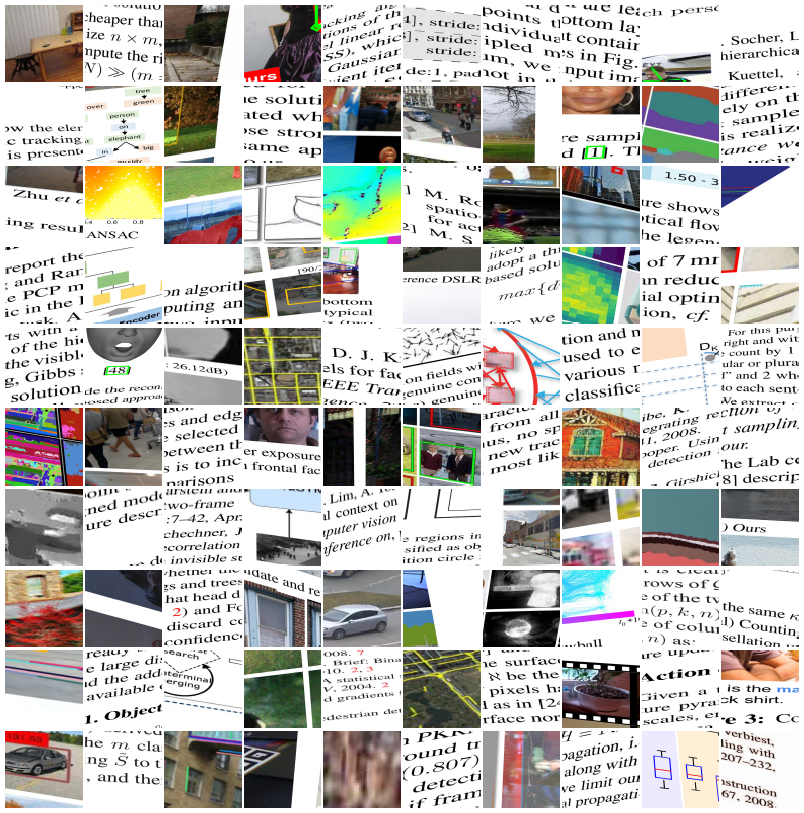} \hspace{0.5mm} &
			\includegraphics[width=0.46\textwidth]{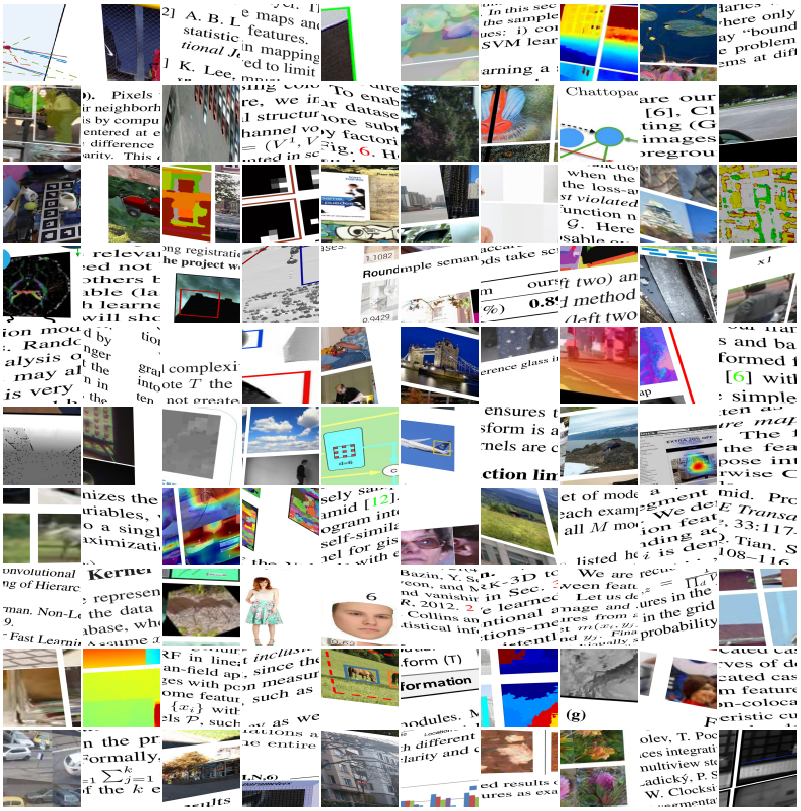}
			\hfill
		\end{tabular}
	
		\caption{Visualization of the proposed \textbf{LCDMoire} validation (left) and test (right) images (moire-free are shown). \textbf{LCDMoire} contains 10000, 100, 100 image pairs for training, validation, test, respectively. Each pair consists of a moire image and a corresponding moire-free image.}
	\label{LCDMoire:valtest}
\end{figure*}

We introduce the LCDMoire dataset, a novel collection of images for benchmarking example-based image demoireing. Figure~\ref{LCDMoire:valtest} gives an overview of the validation and test datasets. 
LCDMoire is intended to complement existing image demoireing datasets~\cite{liu2018demoir, sun2018moire}.

LCDMoire is a synthetically generated dataset that models the increasingly common use case where a user takes a picture of an LCD screen, which may be displaying text and/or pictures. 
The dataset consists of 10,000 training, 100 validation, and 100 test RGB image pairs. Each image has a resolution of $1024 \times 1024$ pixels. The clean images are extracted from existing computer vision conference papers (e.g., papers from ICCV, ECCV, and CVPR). By design, the clean images consist of images of text, figures or combined text/figures in equal proportion. The images are of high quality in terms of the underlying image content, and have varying moire patterns, noise, and compression artifacts.

\subsection{Data generation pipeline}
\label{sec:datagen}

The moire images are generated through a pipeline, similar to~\cite{liu2018demoir}, modeling the picture taking process when a smartphone is used to take pictures from an LCD screen displaying a clean image. 

\begin{figure*}[thbp]
	\centering
	\includegraphics[width=2.0\columnwidth]{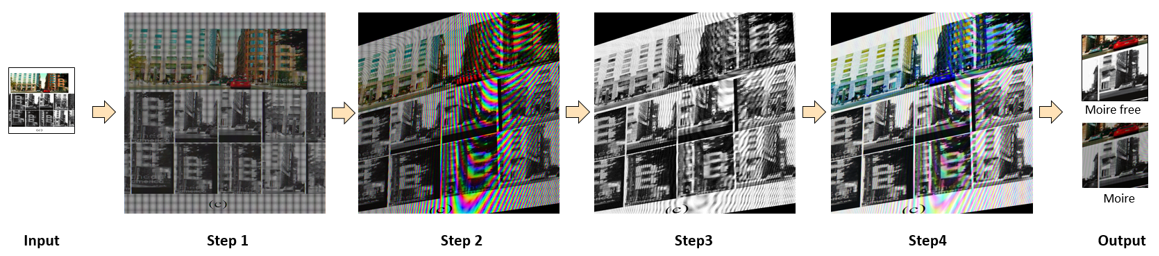}
	\caption{Data generation pipeline.}
	\label{LCDMoire:synthetic}
\end{figure*}

The pipeline, shown in Figure \ref{LCDMoire:synthetic}, includes the following steps:

\begin{enumerate}
    \item Resample the input RGB image into a mosaic of RGB subpixels (modeled as 9 pixels with [K, K, K; R, G, B; R, G, B], where K stands for black) to simulate the image displayed on the LCD. Note that this step causes the final moire image to be darker.
    \item Apply a random projective transformation on the image to simulate different relative positions and orientations of the display and camera, and apply $3 \times 3$ Gaussian filter. 
    To model the random projective transformation, four image corners were randomly sampled (using a uniform distribution) within a radius of 0.2 times of the image size. 
    \item Resample the image using Bayer CFA to simulate the RAW data, and add Gaussian noise to simulate sensor noise.
    \item Apply a simple ISP pipeline including demosaicing (bilinear interpolation) and denoising (standard denoising function provided by OpenCV).
    \item Compress the image using JPEG compression to simulate the compression noise, align the clean image with the moire image, and crop out the image pair. 

\end{enumerate}

\subsection{Post-processing}

\begin{figure}[htbp]
	\centering
	\includegraphics[width=0.90\columnwidth]{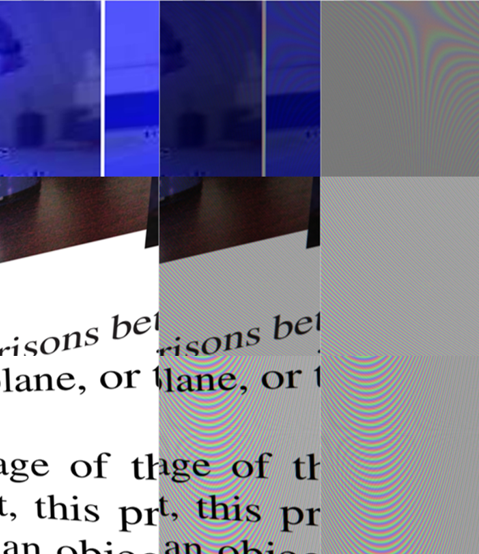}
	\caption{LCD screen examples. Left: clean images, middle: moire images, right: corresponding moire patterns.}
	\label{LCDMoire:examples}
\end{figure}

To ensure the same distributions for the content and moire patterns in the training, validation and testing image sets, some post-processing was conducted:

1) \textbf{Balanced image content.} We wanted to ensure that the dataset was balanced in the sense of having evenly distributed images of text, figures or mixed text / figures. Figure~\ref{LCDMoire:examples} shows examples, where the top row has only figures, the bottom row has only text, and the middle row has both figures and text. For a given image, classification into these three categories was achieved simply based on counting the number of pure white pixels (pixels from the background) and pure black pixels (text pixels).  When the clean image contains only text, the number of pure white and pure black pixels (across the three color channels) together is close to 3145728 ($3145728 = 1024 \times 1024 \times 3$). When the clean image has only figures, then the count is close to 0. Based on this simple classifier, we identified images in equal proportions across these three classes to form the dataset.
In our implementation, the thresholds of 75\% and 25\% of the total number of pixels (3145728) were used to divide the three classes.

\begin{figure}[htbp]
	\centering
	\includegraphics[width=1.0\columnwidth]{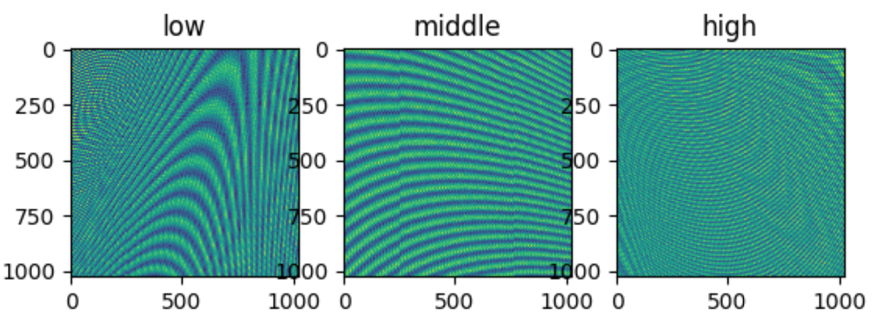}
	\caption{Moire pattern groups. Left: low frequency, middle: middle frequency, right: high frequency.}
	\label{LCDMoire:moirepatterns}
\end{figure}

2) \textbf{Balanced moire frequency content.}	We analyzed the distributions of the moire patterns to ensure a balanced power spectral distribution in the dataset, as shown in  Figure~\ref{LCDMoire:moirepatterns}. We have three groups, namely the low frequency group (Figure~\ref{LCDMoire:moirepatterns} left), middle frequency group (Figure~\ref{LCDMoire:moirepatterns} middle), and high frequency group (Figure~\ref{LCDMoire:moirepatterns} right), based on which frequency group is dominant.
In addition to the RGB image going through the moire image generation pipeline, we put a pure white image though the same pipeline (i.e., the same projective transformation, noise, etc), to produce an image with only moire patterns, see Figure~\ref{LCDMoire:examples} (right column). The power spectral density was analyzed on moire pattern-only image. 

\section{The Challenge}
\label{report:challenge}

\begin{table*}[th]
\normalsize 
  \centering
  \resizebox{2.\columnwidth}{!}{
  \begin{tabular}{lccccccc}
  \toprule 
  \bf Team & \bf User name & \bf PSNR  & \bf SSIM & \bf Time & \bf GPU & \bf Platform & \bf Loss \\ 
  \midrule 

Islab-zju
& \emph{q19911124}
& 44.07
& 0.99 
& 0.25
& 2080Ti
& Tensorflow 
& L1 + L1 sobel\\

MoePhoto \cite{cheng2019multi}
& \emph{opteroncx}
&41.84
& 0.98
& 0.10
& Titan V
& Pytorch 
& L1 charbonnier loss\\

KU-CVIP
& \emph{GibsonGirl}
& 39.54
& 0.97
& 0.07
& Titan X Pascal
& Pytorch 
& L1\\

XMU-VIPLab
&\emph{xdhm2017}
& 39.35 
& 0.98
& 0.13
& RTX 2080
& Pytorch
& L1 + perceptual loss + angular loss\\

IAIR
& \emph{wangjie}
& 38.54 
& 0.98
& 0.43
& 1080Ti 
& Pytorch 
& L2 + perceptual loss\\

PCALab
&\emph{Hanzy}
& 33.72
& 0.98
& 0.02
& 2080Ti 
& Pytorch 
& L1 + contextual loss\\

IPCV\_IITM
&\emph{kuldeeppurohit3}
& 32.23
& 0.96
& -
& Titan X 
& Pytorch
& -\\

Neuro
&\emph{souryadipta}
& 29.87
& 0.92
& 0.25
& 1080Ti
& Pytorch
& -\\

  \bottomrule
  \end{tabular}}
  
    \caption{Results and rankings of methods submitted to the \textbf{fidelity track}. The methods are ranked by PSNR values, \textit{Time} is for runtime per image and measured in seconds.}
  \label{tab:fidelity} 

\end{table*}

\begin{table*}[ht]
\normalsize 
  \centering
  \resizebox{2.\columnwidth}{!}{
  \begin{tabular}{lccccccc}
  \toprule 
  \bf Team & \bf User name & \bf PSNR  & \bf SSIM & \bf Time &\bf GPU  & \bf MOS & \bf Loss \\ 
  \midrule 

Islab-zju
& \emph{q19911124}
& 44.07
& 0.99 
& 0.25
& 2080Ti
& 241
& L1 + L1 sobel\\

XMU-VIPLab
&\emph{xdhm2017}
& 39.35 
& 0.98
& 0.13
& RTX 2080
& 216
& L1 + perceptual loss + angular loss\\

MoePhoto \cite{cheng2019multi}
&\emph{opteroncx}
& 31.07
& 0.96
& 0.11
& Titan V
& 215
& L1 + Perceptual loss\\

KU-CVIP
& \emph{GibsonGirl}
& 39.54
& 0.97
& 0.07
& Titan X Pascal
& 210
& L1\\

Neuro
&\emph{souryadipta}
& 29.87
& 0.92
& 0.25
& 1080Ti
& 187
& -\\

PCALab
&\emph{Hanzy}
& 30.48
& 0.96
& 0.01
& 2080Ti 
& 183
& L1 loss\\

IPCV\_IITM
&\emph{kuldeeppurohit3}
& 32.23
& 0.96
& -
& Titan X 
& 169
& -\\

  \bottomrule
  \end{tabular}}
  
    \caption{Results and rankings of methods submitted to the \textbf{perceptual} track. The methods are ranked by MOS scores, \textit{Time} is for runtime per image and measured in seconds.}
  \label{tab:perceptual} 

\end{table*}

As stated earlier, the demoireing challenge was hosted jointly with the first \textit{Advances in Image Manipulation (AIM 2019)} workshop held in conjunction with the International Computer Vision Conference (ICCV) 2019, Seoul, Korea.

\subsection{Tracks and Evaluation}

The challenge had two tracks: \textbf{Fidelity} and \textbf{Perceptual}, to assess methods in terms of objective and subjective quality.

\textbf{Fidelity:} In this track, the participants developed demoire methods to achieve a moire-free image with the best fidelity compared to the ground truth moire-free image. We used the standard Peak Signal To Noise Ratio (PSNR) and, additionally, the Structural Similarity (SSIM) index as often employed in the literature. The implementations are found in most image processing toolboxes. For each dataset we report the average results over all the processed images.

\textbf{Perceptual:} In this track, the participants developed demoire methods to achieve a moire-free image with the best perceptual quality (measured by Mean Opinion Score (MOS)) when compared to the ground truth moire-free image. In the Mean Opinion Score, human subjects were invited to express their opinions comparing the demoired image to the clean image. Along with the MOS results, we also report the standard Peak Signal To Noise Ratio (PSNR) and the Structural Similarity (SSIM) as a reference.  However, the MOS was the important metric in this track. 

\subsection{Competition}

\textbf{Platform:} The CodaLab platform was used for this competition. To access the data and submit their demoired image results to the CodaLab evaluation server, each participant had to register.

\textbf{Challenge phases:} 
(1) Development (training) phase: The participants received both moire and moire-free training images of the LCDMoire dataset; 
(2) Validation phase: The participants had the opportunity to test their solutions on the moire validation images and to receive immediate feedback by uploading their results to the server. A validation leaderboard was available; 
(3) Final evaluation phase: After the participants received the moire test images and clean validation images, they had to submit both their demoired images and a description of their method before the challenge deadline. One week later the final results were made available to the participants.

\subsection{Challenge results}

From 60 registered participants, eight teams entered in the final phase in the \textbf{fidelity} track, and seven teams in the \textbf{perceptual} track. The teams submitted their results, codes/executables, and factsheets. Table~\ref{tab:fidelity} and Table~\ref{tab:perceptual} report the final test results, rankings of the challenge, self-reported runtimes and major details from the factsheets. 

For the \textbf{fidelity} track, the leading entry as shown in Table~\ref{tab:fidelity} was from the Islab-zju team, scoring a PSNR of $44.07$ dB (higher is better). The second and the third place entries were by the teams MoePhoto and KU-CVIP, respectively.

For the \textbf{perceptual} track, the leading entry as shown in Table~\ref{tab:perceptual} was also from the Islab-zju team, scoring a MOS of $241$ (higher is better). The second and the third place entries were by the teams XMU-VIPLab and MoePhoto, respectively.

\section{Image Quality Assessment (IQA)}

\begin{table}[th]
\normalsize 
  \centering
  \resizebox{1.0\columnwidth}{!}{
  \begin{tabular}{lcccccc}
  \toprule 
  \bf \thead{Team \\ \space } & \bf \thead{PSNR \\ T}  & \bf \thead{SSIM \\ T} & \bf \thead{PSNR \\ F}  & \bf \thead{SSIM \\ F} & \bf \thead{PSNR \\ M}  & \bf \thead{SSIM \\ M}  \\ 
  \midrule 
Islab-zju
&46.57%
&0.99%
&40.86%
&0.97%
&44.24%
&0.99\\

MoePhoto

&45.17
&0.99
&38.02
&0.96
&41.82
&0.99\\

KU-CVIP
&41.90
&0.99
&37.21
& 0.95
&39.20
&0.97\\

XMU-VIPLab
&43.55
&0.99
&35.66
&0.96
&38.57
&0.98\\

IAIR
&43.28
&0.99
&34.29
&0.95
&37.82
&0.98\\

PCALab
&42.03
&0.99
&27.63
&0.95
&31.41
&0.98\\

IPCV\_IITM
&35.42
&0.98
&30.10
&0.93
&31.06
&0.95\\

Neuro
&36.02
&0.99
&24.27
&0.84
&29.19
&0.92\\

  \bottomrule
  \end{tabular}}
  
    \caption{Results for the three classes for the \textbf{fidelity track}. T, F, and M stand for Text only, Figures only, and Mix of figures and text, respectively.}
  \label{tab:IQAfidelity} 

\end{table}

\begin{figure*}[th]
 	\centering
 	\includegraphics[width=2.\columnwidth]{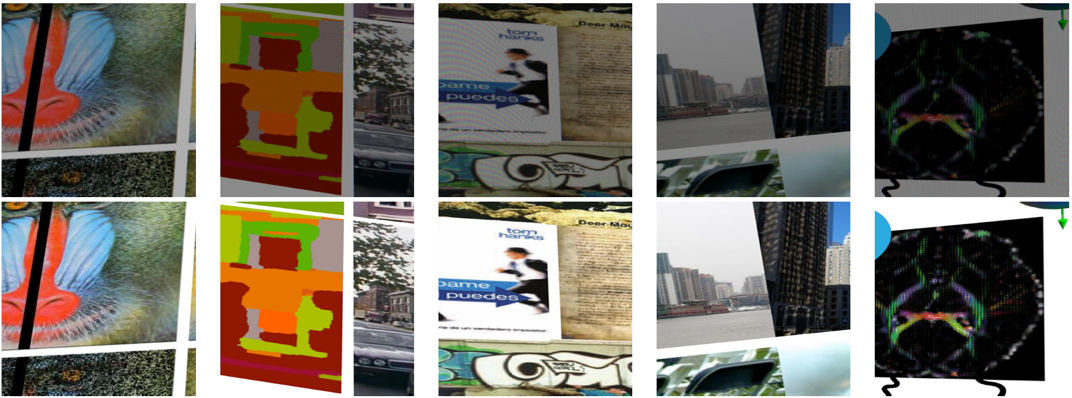}
 	\includegraphics[width=2.\columnwidth]{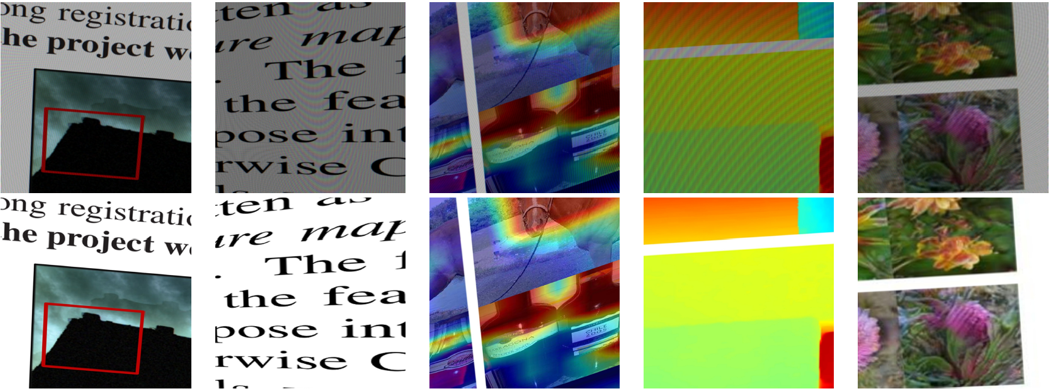}
 	\caption{The ten sampled images used in the MOS study. First and third rows are the moire images, second and fourth rows are the ground truth.}
 	\label{IQA:MOSimages}
 \end{figure*}

\begin{figure}[th]
 	\centering
 	\includegraphics[width=1.\columnwidth]{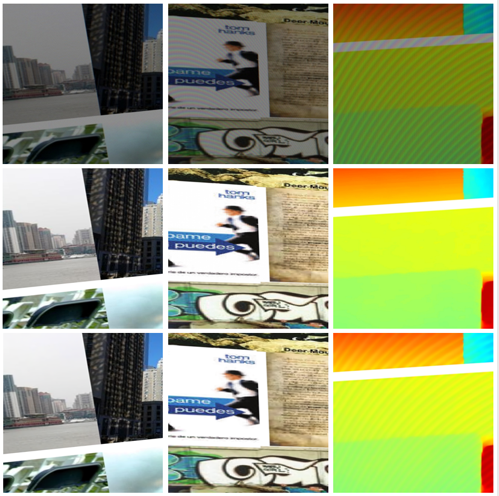}
 	
 	\caption{Example MOS study images. The top row shows the moire images, the middle row contains the ground truth, and the bottom row shows demoired results. From left to right, the results received scores of 3, 4, 1 respectively.}
 	\label{IQA:MOSimages_example}
 \end{figure}

Image quality assessment has received significant attention in the literature with many approaches being proposed.
In this challenge we adopted the standard Peak Signal To Noise Ratio (PSNR) and the Structural Similarity (SSIM) index as often employed in the literature.
PSNR is defined as the pixel-level fidelity relative to the ground truth. 
SSIM 
is a more perceptual quality based model that considers the degradation of images as changes in the perceived structural information.

In the \textbf{fidelity track}, we also evaluated the images for three classes, i.e., Figures, Text and Mix of figures and text. Table \ref{tab:IQAfidelity} shows the PSNR and SSIM for each class. The test set has 33, 33, and 34 images for Figures, Text, and Mix, respectively. The results show that the Text class is the easiest to demoire, followed by Mix and Figures classes.

In the \textbf{perceptual track}, in addition to PSNR and SSIM, we use the Mean Opinion Score (MOS). However, since MOS requires opinions from human subjects it was conducted only on the final test images for the final ranking. 
For the MOS study, we recruited ten judges, including nine professional low-level image processing experts and one IT professional, to rank the submitted results. 
All judges were instructed to use image viewing software FastStone Image Viewer Version 7.4 (FastStone Soft\footnote{https://www.faststone.org}), where they can compare two images (demoired image vs.\ ground truth, or ground truth vs.\ demoired image) side-by-side directly, and can also zoom in to see the paired details very conveniently. 
During the evaluation, the judges did not know which method was used to demoire the image. Also, the relative positioning of the images (ground truth on the left and demoired image on the right, or vice versa) was selected randomly and was unknown to the judges. 

Ranking of all the 100 test images for the seven submitted methods was not possible within a short period of time.
Instead, a subset of 10 representative test images 
(with more focus on the figures-only class to reflect the performance of each method on \textit{hard cases})
were selected for the MOS study. See Figure~\ref{IQA:MOSimages} for the ten test images.
The judges ranked each image for the quality between the reference/ground truth and the demoired image on a scale from 1 to 5 (definitely worse quality (1), probably worse (2), probably similar (3), probably better (4), definitely better (5)). The queries were randomly flipped reference-solution and solution-reference, and the order of the queries was also random. So for each query we had 10 scores in total.

The final MOS score for a method was the accumulated rankings from the ten judges and from the 10 test images. Based on this protocol, a demoire method could receive a score in the range of [100, 500], the higher the better.
There were seven teams that submitted their results for the \textbf{Perceptual} track; see Table~\ref{tab:perceptual} for the rankings. Islab-zju team obtained a score of 241, followed by XMU-VIPLab team (216) and MoePhoto team (215).

In many cases, the judges preferred the ground truth image over the demoired image. However, there are some outliers, where some of the judges preferred the demoired image over the ground truth. Figure~\ref{IQA:MOSimages_example} shows example images used in the MOS study.  
The top row, middle row and bottom row are the moire images, ground truth images, and demoired images, respectively (results from IsLab-zju). During the evaluation, the judges were given the pairs of images (the bottom two rows) to which they assigned the scores.
As an example, the image received a score of 3, meaning that the judge assessed the quality of images (the ground truth and the demoired image) as \textit{probably similar}.
The middle image received a score of 4,  meaning that the judge preferred the quality of the demoired image (\textit{probably better}) over the ground truth. The right image received a score of 1, meaning the judge rated it to have a \textit{definitely worse} quality than the ground truth.

\section{Related datasets}

In this section, we describe the other datasets that appear in image demoireing literature. 

\begin{figure}[t]
	\centering
    	\includegraphics[width=1.0\columnwidth]{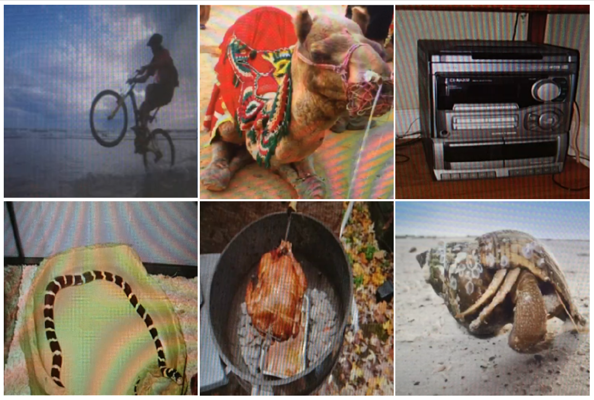}
	\caption{Examples from TIP18~\cite{sun2018moire} dataset.}
	\label{TIP:examples}
\end{figure}

One relevant dataset is TIP2018 dataset~\cite{sun2018moire}, see Figure~\ref{TIP:examples} for examples. 
The dataset has 135,000 real image pairs, each containing a clean image and a moire image. The clean images come from 100,000 validation images and 50,000 testing images of the ImageNet ILSVRC 2012 dataset~\cite{russakovsky2015imagenet}. The corresponding moire images are captured with three smartphones (iPhone 6, SAMSUNG Galaxy S7 Edge, and SONY Xperia Z5 Premium Dual) from three LCD screens (Macbook Pro Retina, DELL U2410 LCD, DELL SE198WFP LCD).

The dataset was created through two steps: capture and alignment.
For capture, the images are manually taken from the screens displaying the clean images, which were enhanced with a special black border. 
For alignment, 20 sharp corners on the black border, from the moire image, were detected using a Harris corner detector~\cite{harris1988combined}, and filtered through heuristics to avoid false detections. Then a homography was used for the alignment.
However the dataset has some alignment issues, originating from the corner detection step. 

\begin{figure}[t]
	\centering
    	\includegraphics[width=1.0\columnwidth]{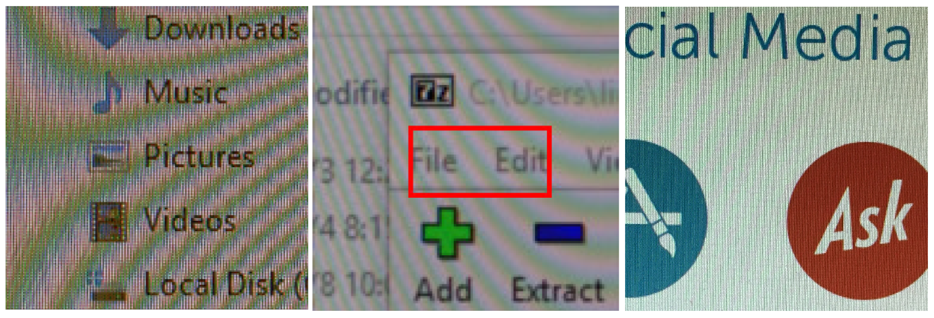}
	\caption{Examples extracted from the arXiv paper~\cite{liu2018demoir}.}
	\label{bolin:examples}
\end{figure}

Another dataset was proposed in \cite{liu2018demoir}, see Figure~\ref{bolin:examples} for examples extracted from the paper. 
The dataset has 80,000 synthetic image pairs of $512 \times 512$ resolution.
The moire images were synthetically created through a similar pipeline to the one used to create LCDMoire. The clean images come from 1,000 digital images with various content commonly displayed on a computer, e.g., dialog box, text, graphics, etc. 
Unfortunately, at the time of hosting the workshop, we could not find any download link, even though the paper was submitted to ArXiv in April 2018.

Therefore, it was necessary to create a new dataset for the purpose of the demoire challenge. Unlike the dataset in~\cite{sun2018moire}, LCDMoire does not suffer from alignment issues. And inspired by the pipeline from~\cite{liu2018demoir}, we created the LCDMoire dataset but made it publicly available to enable comparisons between demoire methods.

\section{Challenge methods}

\subsection{Islab-zju team}
\label{report:q19911124}

The Islab-zju team is the winning team for both tracks.
They proposed a \textit{CNN Based Learnable Multiscale Bandpass Filter method}, dubbed \emph{MBCNN}. 
They referenced the architecture of CAS-CNN~\cite{CAS-CNN} and proposed a 3-level multiscale bandpass convolutional neural network.
MBCNN first uses pixel shuffle to reduce the input image resolution to half.
Subsequent scales are generated using two $2\times2$ stepped convolution layers to start the other two levels. %
For each level, the processing can be divided into three parts: artifact reduction, tone mapping and image reconstruction.

\subsection{MoePhoto team}
\label{report:opteroncx}

The MoePhoto team proposed the method of \textit{Multi Scale Dynamic Feature Encoding Network for Image Demoireing}, dubbed \textit{MDDM} \cite{cheng2019multi}.
This method is inspired by \cite{liu2018demoir, gao2019moire, sun2018moire}.
MDDM constructs an image feature pyramid, encodes image features at different spatial resolutions, and obtains image feature representations of different frequency bands. 
MDDM also proposes a dynamic feature encoding (DFE) method. At each downsampling branch, the method adds an extra lightweight branch.

\subsection{XMU-VIPLab team}
\label{report:xdhm2017}

XMU-VIPLab team proposed \textit{Global-local Fusion Network for Single Image Demoireing}. 
The method first trains a local and a global demoireing network, respectively. It then uses a fusion branch to fuse the results of the two networks to obtain the final demoireing result. 
The local network uses an asymmetric encode-decode architecture, using the Residual in Residual blocks as the basic block.
The global network uses a standard U-Net architecture, and the input is scaled by 4. 
The fusion network only contains 4 residual modules. 
When fusing, the result of the global network needs to be amplified by 4 times.

\subsection{KU-CVIP team}
\label{report:GibsonGirl}

The KU-CVIP team proposed the \textit{Densely-Connected Residual Dense Network}, inspired by \cite{RDN, kim2019grdn}. 
The network is based on Residual Dense Block (RDB)~\cite{RDN}. 
The method connects RDBs densely and adds skip connections in each DCRDB unit. 
The method downsamples and upsamples the feature map using $4 \times 4$ convolution layers with stride 2 to extract multi-scale features and make the network deeper. 
Starting with 1 RDB in a DCRDB unit, the method increases/decreases the number of RDBs in DCRDB by a factor of 2 whenever downsampling/upsampling the feature map. 
At the end of the network, the method applies a CBAM \cite{woo2018cbam} module and $3 \times 3$ convolution layer to make the input and output image have same number of channels.

\subsection{IAIR team}
\label{report:wangjie}

The IAIR team proposed \textit{Wavelet-Based Multi-Scale Network}, namely WMSN, for Image Demoireing.
This network is based on the U-Net~\cite{UNet} architecture, in which the clique wavelet block~\cite{nips18} was used in the decoder part to restore the degraded features and upsample the restoration features.
The proposed network contains four strided convolutional layers and four wavelet blocks. 
The first four CABlocks in encoder part consists of three residual channel attention blocks~(RCAB) by~\cite{rcan}, and 18 RCABs are used in the last CABlocks. 
The clique wavelet blocks are used to remove the moire artifacts from the features in the wavelet domain and upsample the restoration features.

\subsection{PCALab team}
\label{report:hanzy}

The PCALab team proposed \textit{Domain Adaptation for Image Demoireing}. The method is inspired by~\cite{tzeng2017adversarial, mechrez2018contextual}.
The method proposes to remove moire pattern through a domain adaptation method, in which the method assumes that the clean images come from clean domain and the images with moire pattern are from a moire domain and the two domains share the same latent space. The method incorporates an Auto Encoder (AE) framework, where the encoder is designed to extract the common latent features shared by two domains and the decoder restores the images from corresponding latent features.
In the \textbf{fidelity} track, the loss is L1 loss + contextual loss.
In the \textbf{perceptual} track, the contextual loss is removed.

\subsection{Neuro team}
\label{report:souryadipta}

The Neuro team proposed a method titled \textit{Demoireing High Resolution Images Using Conditional GANs}.
This method models the problem as a image translation problem and uses conditional GANs to reconstruct the demoired image, inspired by~\cite{wang2018high}. 
Reconstructing certain frequency components from rest of the frequency spectrum of a image is the main goal for this solution. 
The method employs a Global Generator instead of local Generator to emphasize global features more to increase fidelity and the perceptual quality of the restored image.

\subsection{IPCV\_IITM team}
\label{report:kuldeeppurohit3}

The IPCV\_IITM team proposed a \textit{Sub-pixel Dense U-net for Image Demoireing}. %
Following \cite{2019arXiv190311394P}, the proposed network consists of a deep dense-residual encoder-decoder structure with subpixel convolution and multi-level pyramid pooling module for efficiently estimating the demoireing image.
The resolution of input image is brought down by a factor of 2 using pixel-shuffling before being fed to the encoder, which increases the receptive field of the network and reduces computational footprint. The encoder is made of densely connected modules.

\section{Conclusion}

In this paper, we introduced LCDMoire, a new dataset for benchmarking image demoireing methods. 
We provide high-quality ground truth reference frames as well
as corresponding moire frames. 
Each degraded frame
models commonly occurring video degradations such as
motion blur, compression, and downsampling. 
We studied the submitted solutions from the AIM 2019 image demoireing challenge.

The LCDMoire dataset proved an effective dataset for the challenge, enabling participants to develop and evaluate deep learning demoire approaches.  The dataset was sufficiently challenging to produce differentiated results between submitted solutions.  

The dataset does have some limitations that could be addressed in future work.  First, we assumed a single LCD pattern (consisting of nine subpixels described in Section~\ref{sec:datagen}.  In practice, more LCD subpixel layouts are possible.  Additionally, we assumed the common Bayer pattern CFA, but other CFA designs are well known in the literature.  Despite these limitations, LCDMoire captures a wide variety of moire patterns.  Nonetheless, there likely exists a domain gap between the synthesized moire patterns of LCDMoire and moire patterns that exist in photographs taken with specific cameras.  Future work could study this domain gap.  

We hope the LCDMoire dataset will inspire further demoire research and be used to benchmark performance of future single image demoire methods.

{\small
\bibliographystyle{ieee_fullname}
\bibliography{egbib}

\begin{thebibliography}{10}\itemsep=-1pt

\bibitem{CAS-CNN}
Lukas Cavigelli, Pascal Hager, and Luca Benini.
\newblock {CAS}-{CNN}: A deep convolutional neural network for image
  compression artifact suppression.
\newblock In {\em 2017 International Joint Conference on Neural Networks
  ({IJCNN})}. {IEEE}, may 2017.

\bibitem{cheng2019multi}
Xi Cheng, Zhenyong Fu, and Jian Yang.
\newblock Multi-scale dynamic feature encoding network for image
  demoir{\'e}ing.
\newblock In {\em The IEEE International Conference on Computer Vision (ICCV)
  Workshops}, 2019.

\bibitem{gao2019moire}
Tianyu Gao, Yanqing Guo, Xin Zheng, Qianyu Wang, and Xiangyang Luo.
\newblock Moir{\'e} pattern removal with multi-scale feature enhancing network.
\newblock In {\em 2019 IEEE International Conference on Multimedia \& Expo
  Workshops (ICMEW)}, pages 240--245. IEEE, 2019.

\bibitem{harris1988combined}
Christopher~G Harris, Mike Stephens, et~al.
\newblock A combined corner and edge detector.
\newblock In {\em Alvey vision conference}, volume~15, pages 10--5244.
  Citeseer, 1988.

\bibitem{kim2019grdn}
Dong-Wook Kim, Jae Ryun~Chung, and Seung-Won Jung.
\newblock Grdn: Grouped residual dense network for real image denoising and
  gan-based real-world noise modeling.
\newblock In {\em Proceedings of the IEEE Conference on Computer Vision and
  Pattern Recognition Workshops}, pages 0--0, 2019.

\bibitem{liu2018demoir}
Bolin Liu, Xiao Shu, and Xiaolin Wu.
\newblock Demoireing of camera-captured screen images using deep convolutional
  neural network.
\newblock {\em arXiv preprint arXiv:1804.03809}, 2018.

\bibitem{mechrez2018contextual}
Roey Mechrez, Itamar Talmi, and Lihi Zelnik-Manor.
\newblock The contextual loss for image transformation with non-aligned data.
\newblock In {\em Proceedings of the European Conference on Computer Vision
  (ECCV)}, pages 768--783, 2018.

\bibitem{2019arXiv190311394P}
Kuldeep {Purohit} and A.~N. {Rajagopalan}.
\newblock {Region-Adaptive Dense Network for Efficient Motion Deblurring}.
\newblock {\em arXiv e-prints}, page arXiv:1903.11394, Mar 2019.

\bibitem{UNet}
Olaf Ronneberger, Philipp Fischer, and Thomas Brox.
\newblock U-net: Convolutional networks for biomedical image segmentation.
\newblock In {\em International Conference on Medical Image Computing and
  Computer-assisted Intervention}, 2015.

\bibitem{russakovsky2015imagenet}
Olga Russakovsky, Jia Deng, Hao Su, Jonathan Krause, Sanjeev Satheesh, Sean Ma,
  Zhiheng Huang, Andrej Karpathy, Aditya Khosla, Michael Bernstein, et~al.
\newblock Imagenet large scale visual recognition challenge.
\newblock {\em International journal of computer vision}, 115(3):211--252,
  2015.

\bibitem{sun2018moire}
Yujing Sun, Yizhou Yu, and Wenping Wang.
\newblock Moire photo restoration using multiresolution convolutional neural
  networks.
\newblock {\em IEEE Transactions on Image Processing}, 27(8):4160--4172, 2018.

\bibitem{tzeng2017adversarial}
Eric Tzeng, Judy Hoffman, Kate Saenko, and Trevor Darrell.
\newblock Adversarial discriminative domain adaptation.
\newblock In {\em Proceedings of the IEEE Conference on Computer Vision and
  Pattern Recognition}, pages 7167--7176, 2017.

\bibitem{wang2018high}
Ting-Chun Wang, Ming-Yu Liu, Jun-Yan Zhu, Andrew Tao, Jan Kautz, and Bryan
  Catanzaro.
\newblock High-resolution image synthesis and semantic manipulation with
  conditional gans.
\newblock In {\em Proceedings of the IEEE conference on computer vision and
  pattern recognition}, pages 8798--8807, 2018.

\bibitem{woo2018cbam}
Sanghyun Woo, Jongchan Park, Joon-Young Lee, and In So~Kweon.
\newblock Cbam: Convolutional block attention module.
\newblock In {\em Proceedings of the European Conference on Computer Vision
  (ECCV)}, pages 3--19, 2018.

\bibitem{AIM19demoireMethods}
Shanxin Yuan, Radu Timofte, Gregory Slabaugh, Ales Leonardis, Bolun Zheng, Xin
  Ye, Xiang Tian, Yaowu Chen, Xi Cheng, Zhenyong Fu, Jian Yang, Ming Hong,
  Wenying Lin, Wenjin Yang, Yanyun Qu, Hong-Kyu Shin, Joon-Yeon Kim, Sung-Jea
  Ko, Hang Dong, Yu Guo, Jie Wang, Xuan Ding, Zongyan Han, Sourya~Dipta Das,
  Kuldeep Purohit, Praveen Kandula, Maitreya Suin, and Rajagoapalan A.N.
\newblock Aim 2019 challenge on image demoreing: methods and results.
\newblock In {\em The IEEE International Conference on Computer Vision (ICCV)
  Workshops}, 2019.

\bibitem{rcan}
Yulun Zhang, Kunpeng Li, Kai Li, Lichen Wang, Bineng Zhong, and Yun Fu.
\newblock Image super-resolution using very deep residual channel attention
  networks.
\newblock In {\em European Conference on Computer Vision}, pages 286--301,
  2018.

\bibitem{RDN}
Yulun Zhang, Yapeng Tian, Yu Kong, Bineng Zhong, and Yun Fu.
\newblock Residual dense network for image super-resolution.
\newblock In {\em 2018 {IEEE}/{CVF} Conference on Computer Vision and Pattern
  Recognition}. {IEEE}, jun 2018.

\bibitem{nips18}
Zhisheng Zhong, Tiancheng Shen, Yibo Yang, Zhouchen Lin, and Chao Zhang.
\newblock Joint sub-bands learning with clique structures for wavelet domain
  super-resolution.
\newblock In {\em Neural Information Processing Systems}, pages 165--175, 2018.

\end{thebibliography}
}

\end{document}